\documentclass{article}
\usepackage[preprint]{spconf}

\usepackage{color,xcolor}
\usepackage{epsfig}
\usepackage{graphicx}

\usepackage{array}
\usepackage{booktabs}
\usepackage{colortbl}
\usepackage{multirow}
\usepackage{float}
\usepackage{footnote}
\makesavenoteenv{tabular}
\makesavenoteenv{table}

\usepackage{amsmath,amsfonts,amssymb,bm}
\usepackage[super]{nth}

\usepackage{changepage}
\usepackage{extramarks}
\usepackage{fancyhdr}
\usepackage{lastpage}
\usepackage{setspace}
\usepackage{soul}
\usepackage{xspace}

\usepackage{url}
\usepackage{hyperref}
\hypersetup{colorlinks=True, urlcolor=black}
\usepackage[numbers,sort&compress]{natbib}

\usepackage{algorithm, algorithmic}
\usepackage{enumitem}
\usepackage{verbatim}
\usepackage{pifont}
\usepackage[acronyms]{glossaries}
\glsdisablehyper
\newcommand{\sect}[1]{Section~\ref{sec:#1}}
\newcommand{\sectdot}[1]{Sec.~\ref{sec:#1}}

\newcommand{\tbl}[1]{Table~\ref{tab:#1}}

\newcommand{\twosect}[2]{Sections~\ref{sec:#1} and \ref{sec:#2}}

\newcommand{\ignore}[1]{}

\makeatletter
\DeclareRobustCommand\onedot{\futurelet\@let@token\@onedot}
\def\@onedot{\ifx\@let@token.\else.\null\fi\xspace}

\def\eg{\emph{e.g}\onedot} 
\def\ie{\emph{i.e}\onedot}

\makeatother

\definecolor{MyDarkBlue}{rgb}{0,0.08,1}
\definecolor{MyDarkGreen}{rgb}{0.02,0.6,0.02}
\definecolor{MyDarkRed}{rgb}{0.8,0.02,0.02}
\definecolor{MyDarkOrange}{rgb}{0.40,0.2,0.02}
\definecolor{MyPurple}{RGB}{111,0,255}
\definecolor{MyRed}{rgb}{1.0,0.0,0.0}
\definecolor{MyGold}{rgb}{0.75,0.6,0.12}
\definecolor{MyDarkgray}{rgb}{0.66, 0.66, 0.66}

\def\presec{\vspace{-0.25em}}
\def\postsec{\vspace{-0.15em}}

\definecolor{spkA}{HTML}{1f77b4}
\definecolor{spkB}{HTML}{ff7f0e}
\definecolor{spkC}{HTML}{2ca02c}
\definecolor{spkD}{HTML}{d62728}
\definecolor{spkE}{HTML}{9467bd}

\newacronym{rnn}{RNN}{recurrent neural network}
\newacronym{ann}{ANN}{artificial neural network}
\newacronym{mlp}{MLP}{Multi-layer perceptron}
\newacronym{asr}{ASR}{automatic speech recognition}
\newacronym{ssl}{SSL}{self-supervised learning}
\newacronym{mppt}{MPPT}{masked prediction pre-training}
\newacronym{wer}{WER}{word error rate}
\newacronym{hubert}{HuBERT}{Hidden-Unit BERT}
\newacronym{ctc}{CTC}{Connectionist Temporal Classification} %
\newacronym{bpe}{BPE}{Byte Pair Encoding} %
\newacronym{ce}{CE}{cross-entropy}
\usepackage{spconf,amsmath,graphicx}
\newcolumntype{H}{>{\setbox0=\hbox\bgroup}c<{\egroup}@{}}

\title{Biased Self-supervised learning for ASR}
\name{\begin{tabular}{c}
Florian L. Kreyssig$^{\star,1}$, Yangyang Shi$^2$, Jinxi Guo$^2$, Leda Sari$^2$,\\
Abdelrahman Mohamed$^2$, Philip C. Woodland$^1$ \thanks{$^\star$Work done while Florian Kreyssig was an Intern at Meta AI.}%
\end{tabular}}
\address{$^1$University of Cambridge, UK, $^2$Meta AI, USA\\\small{\texttt{flk24@cam.ac.uk\vspace{-0.5em}}}}
\begin{document}
\ninept
%
\maketitle
\begin{abstract}
Self-supervised learning via \gls{mppt} has shown impressive performance on a range of speech-processing tasks.
This paper proposes a method to bias self-supervised learning towards a specific task. The core idea is to slightly finetune the model that is used to obtain the target sequence.
This leads to better performance and a substantial increase in training speed.
Furthermore, this paper proposes a variant of \gls{mppt} that allows low-footprint streaming models to be trained effectively by computing the \gls{mppt} loss on masked and unmasked frames.
These approaches are evaluated for \acrlong{asr} on the Librispeech corpus, where 100 hours of data served as the labelled data and 860 hours as the unlabelled data. The biased training outperforms the unbiased training by 15.5\% after 250k updates and 23.8\% after 100k updates on test-other. For the streaming models, the pre-training approach yields a reduction in \acrlong{wer} of 44.1\%.

\end{abstract}
\presec
\section{Introduction}
\postsec
\label{sec:intro}
\Gls{ssl} for speech and audio~\cite{mohamed2022ssl} learns representations of audio by being trained to predict targets that are derived from the input audio itself. Such tasks include learning to predict multiple feature encodings of the input~\cite{Pascual2019SSLSpeech,ravanelli2020multiSSLspeech}, distinguishing near-by future speech features from temporally distant ones~\cite{oord2018representation, Schneider2019wav2vec} or directly predicting future frames~\cite{Chung2019APC}, and prediction of audio features underneath a mask~\cite{baevski2019vq,baevski2020wav2vec,hsu2021huberticassp,hsu2021hubert}. These self-supervised algorithms are able to leverage large amounts of unlabelled data effectively.

In \gls{asr}, after a model is pre-trained using one of these algorithms on the unlabelled data, the model is finetuned using labelled data.
Usually, the amount of labelled data is much smaller than the amount of unlabelled data.
It is this small amount of labelled data that we will use to bias our self-supervised learning towards \gls{asr}.

One of the most successful self-supervised learning algorithms for speech has been \gls{hubert}~\cite{hsu2021hubert}. \gls{hubert} clusters speech frames in order to obtain a tokenisation of the input, which is used as the label sequence. During training, portions of the input are masked, and the neural network encoder is trained to predict the corresponding tokens that lie underneath the masks. This style of training will be referred to as \acrfull{mppt}.
For the second iteration of training, hidden embeddings of a model after the first iteration of \gls{mppt} are clustered to obtain an improved tokenisation.
In order to model the evolution of the speech frames, as needed in the first iteration of training, the model has to learn and represent the underlying phones. 
Hence, the tokenisation that is used for the second iteration of training has a strong correlation with the underlying phones~\cite{hsu2021hubert}.

Although \gls{hubert} demonstrates competitive \glspl{wer} after finetuning, there are some problems that cannot be overcome without any supervised signal. \gls{asr} performance is not only about obtaining a concise representation of the audio but also about learning which information to ignore (\eg speaker or channel information). The cluster tokens that are obtained for training can represent various phones. However, they represent phones in various contexts in terms of the channel and the speaker. This leads to the \gls{mppt} task being much more difficult than it would otherwise need to be, which, in turn, leads to slow training and needing many updates (\cite{hsu2021hubert, wang2022supervision} train for 400k updates, \cite{chen2022wavlm} trains for 1m updates).

To address this issue, this paper proposes a method to bias the \gls{hubert} masked prediction task towards \gls{asr} using the labelled data used in the finetuning stage. To bias the \gls{mppt} towards \gls{asr}, the model used to obtain the tokenisation used in the second iteration of training is slightly finetuned using an \gls{asr}-loss. This leads to a tokenisation that is specialised for \gls{asr} and is also more consistent and predictable.
The results will show that this leads to much faster training.
Furthermore, by treating the biased tokens as discovered phonetic units, models can be pre-trained by predicting these tokens directly, even for unmasked audio. This paper proposes that this style of training, which is akin to cross-entropy training of DNN-HMM hybrid models~\cite{dahl2012context}, can be highly effective for training streaming \gls{asr} models with a restricted input window.

The remainder of this paper is organised as follows. \sect{method} outlines the biased semi-supervised learning method, a variant of \gls{mppt} for low-footprint streaming models and discusses related work. \twosect{expsetup}{expres} present the
experimental setup and results and \sect{conclusions} gives conclusions.
\presec
\section{Method}
\postsec
\label{sec:method}

\subsection{HuBERT}\label{sec:hubert}
\postsec
This section summarises~\cite{hsu2021hubert}, the model structure, how to obtain the label sequences and how to train the model on this label sequence.

In~\cite{hsu2021hubert}, the model structure uses a convolutional encoder to encode the raw waveform input. 
This is followed by a convolutional layer with a large kernel size that is used for positional encodings. 
This is followed by multiple (\eg 12 in HuBERT-Base) Transformer layers~\cite{vaswani2017attention}. The output embeddings from the final Transformer layer are then used to predict the underlying token label. This paper uses a modified and more efficient model structure that is explained in detail in \sect{model}.

To train the model, a label sequence is obtained by clustering MFCC features or hidden embeddings. Clustering is performed globally by using the entire corpus (or a representative sample of it) and not locally for each sequence. The model is then trained by first masking portions of the output of the convolutional encoder and then predicting the labels that lie underneath those masks. 

\gls{hubert} training occurs in multiple iterations. The first iteration of training uses a target sequence obtained by clustering speech features using KMeans. For the succeeding iterations, hidden embeddings from the trained model of the previous iteration are clustered. The correct layer to use for this can be found based on the metrics described in \sect{metrics}.

Details about the exact training parameters for the baseline models used in this paper are given in \sect{baselinepipeline}. This training protocol will be referred to as \emph{unbiased}.

\presec
\subsection{Biased Masked Prediction Pre-Training}
\postsec
Clustering is challenging and inherently ambiguous when samples of different clusters are not well separated in the feature space. Without a supervision signal, clustering is not well-defined.
This paper introduces a supervised signal into the pipeline of \sect{hubert} through a small but important change. The first iteration of training is identical to the unbiased \gls{mppt} training. 
The difference to the unbiased \gls{mppt} training comes before the second iteration of \gls{mppt}.
To bias the \gls{mppt} towards the desired task, the trained model of the first iteration is finetuned using a supervised loss for a small number of updates. This biases the model and, thus, the clusters towards the supervised task. Whilst this paper uses a \gls{ctc}~\cite{Graves2006CTC} loss to make the \gls{mppt} \gls{asr} specific, this biasing step could be done using any speech processing task or even multi-task training. After the biasing step, the embeddings are clustered using KMeans clustering. 
Our results will later show that compared to clustering unbiased embeddings, a far smaller number of clusters is necessary and even advantageous. More detail about the exact training parameters is given in \sect{biasedpipeline}. This training protocol will be referred to as \emph{biased}.

In self-training~\cite{zavaliagkosUtilizingUntranscribedTraining1998,kempUNSUPERVISEDTRAININGSPEECH1998,wesselUnsupervisedTrainingAcoustic2001,lamelUnsupervisedAcousticModel2002,park2020improved} supervised data is used to build a seed model, which is then used to label the unlabelled data. The final model is then trained using the combined data. Whilst these methods have a similar operating point to our approach, the difference is that these methods are bootstrapped from a fully supervised model, which can easily overfit the initial data set. Our method is biased to the supervised set using a small number of updates to avoid any overfitting. Furthermore, self-training methods rely on a strong language model as shown by~\cite{wallington21lm4ssl}. Our method does not rely on a language model for pre-training the encoder. \cite{wang2022supervision} uses self-training in an \gls{mppt} setup where the tokenisation is obtained using a seed hybrid or \gls{ctc} system.

\presec
\subsection{Low-footprint streaming models}\label{sec:lowfootprint}
\postsec
\gls{mppt} is challenging for streaming models that have small input contexts. The reason for this is that too much of the input is masked in order to make accurate predictions. Making the task prohibitively difficult leads to bad representations. On top of this, removing the 1D-Convolutional layer further improves efficiency but would make the \gls{mppt} task even harder. This paper solves this issue by computing the classification loss on both the masked and the unmasked frames. By treating the clusters obtained from clustering the embeddings of a second-iteration biased model as acoustic units, the loss obtained from the unmasked frames can be treated as \gls{ce} training for DNN-HMM hybrid models~\cite{dahl2012context,bourlard1994connectionist}. In practice, the input was masked as before, but the loss-weight on the masked frames and the loss-weight on the unmasked frames are both 0.5. Most of the learning comes from the unmasked frames, and the masking serves as a form of regularisation.
\cite{doutre2021improving, yang2022knowledge} train a streaming model from a non-streaming model using teacher-student training. The teacher in \cite{yang2022knowledge} is initialised from a self-supervised model.

\presec
\subsection{Additional Metrics}
\postsec
\label{sec:metrics}
The training and evaluation pipeline, including pre-training, finetuning and decoding, has many stages.
Hence, the final \gls{wer} should not be the only metric to analyse the examined approaches.

\presec
\paragraph*{Masked Prediction Accuracy}
This is the accuracy that the model achieves on the \gls{mppt} task. For the same label sequence, this metric is correlated with final \gls{wer}. For the same model architecture and training steps, this metric is an indication of how ``easy" the task is. The task can be made easier simply by reducing the number of clusters K. It can also be easier because the tokenisation is more consistent. This increased consistency can be viewed to be analogous to having less label noise in supervised training. Thus, this is desirable.

\presec
\paragraph*{Cluster-Purity and Label-purity}
These two metrics are calculated based on labelling each frame with its corresponding cluster and with a ground-truth label. In this work, the data is aligned with 870 clustered bi-character units. 
Cluster-purity is the maximum achievable accuracy of predicting the cluster purely based on the label. 
Label-purity is the maximum achievable accuracy of predicting the ground-truth label purely based on the cluster. These metrics indicate how much the cluster tokens are correlated with the ground-truth labels, and therefore they give an indication towards how similar the \gls{mppt} is to \gls{asr} training. Hence, these metrics are highly correlated with final decoding \gls{wer}. It is important to note that when increasing K for the same embeddings, the cluster-purity will go down, and the label-purity will go up. Therefore these statistics are difficult to compare for models with different K. 
\presec
\section{Experimental Setup}
\label{sec:expsetup}
\subsection{Data}
\postsec
The proposed methods were evaluated on the Librispeech data~\cite{panayotov2015librispeech}. It contains 960 hours of data from audiobooks. A random 100-hour subset serves as the supervised data set that is used for biasing and finetuning. The other 860 hour serves as the unsupervised data set. For the \gls{mppt}, the whole 960 hours of data is used for training. The utterances in Librispeech were segmented to be of length $\leq\!10s$.
\presec
\subsection{Model}
\postsec
\label{sec:model}
The model structure used in this paper has several differences from the one used in~\cite{hsu2021hubert}. The convolutional encoder with an output frequency of 50 Hz, which processes the raw-waveform input, is replaced with a conventional {80dim} FBANK analysis operating at 100~Hz with a 25~ms window size.
The size of the convolutional kernel of the 1D-Convolution layer used for relative positional encodings is reduced from 127 (at 50~Hz, equiv. to 2.54 seconds) down to 31 (at 100~Hz, equiv. to 0.31 seconds). This layer is followed by stacking 4 frames which reduces the frequency of the encoder outputs down to 25~Hz. Furthermore, the Transformer layers~\cite{vaswani2017attention} are replaced by the much more efficient Emformer layers~\cite{shi2021emformer}. The encoder has 20 Emformer layers with an embedding dimension of 512, a feed-forward dimension of 2048, and 8 attention heads. For most experiments, the Emformer layers use a right-context of 1 and a centre-context of 160. For the low-footprint streaming models (see \sect{lowfootprint}), a centre-context of 4 was used, equivalent to 160~ms, and the 1D-Convolution layer was removed. This yields an algorithmic latency induced by the encoder of 120~ms.
\presec
\subsection{Notation}
\postsec
To make this paper more readable, models pre-trained with \gls{mppt} are given a model ID $X^{Y}_{Z}$. $X$ relates to the vectors that are clustered to obtain the label-sequence for training, \eg $M$ for 39dim-MFCC features. $Y$ relates to how long the model was trained \eg 100k for 100k updates. $Z$ relates to the number of clusters used, \eg 100. $M^{100k}_{100}$ is a model trained for 100k updates for which the training labels are obtained by clustering 39dim-MFCC features using K=100.

\presec
\subsection{Baseline Pipeline}
\postsec
\label{sec:baselinepipeline}
This pipeline is aligned with \gls{hubert}-Base~\cite{hsu2021hubert} and is entirely unsupervised. The first iteration of \gls{mppt} uses a label sequence obtained by clustering 39dim-MFCC\footnote{13dim plus delta and deltadelta.} features using K=100. These features are at 100~Hz, and the label sequence is sub-sampled by a factor of 4 to align with the 25~Hz output frequency of the model. The models that are trained for 100k and 250k updates have the IDs $M^{100k}_{100}$ and $M^{250k}_{100}$, respectively and will also be referred to as iteration-1 models. MFCC features are used over FBANK features due to their better match with the feature independence assumption of the squared distance metric of the KMeans algorithm. 
The second iteration of training uses a label sequence obtained by clustering the embeddings of one of the layers of $M^{250k}_{100}$ using K=500.
For this second clustering step, a layer that is in the middle of the model should be used, \ie if 12 layers are used, use layer 6. For this paper's 20-layer model, layer 10 was used. The models that are trained for 100k, 250k and 400k updates have the IDs $H^{100k}_{500}$, $H^{250k}_{500}$ and $H^{400k}_{500}$, respectively. These models will also be referred to as iteration-2 models.


\presec
\subsection{Biased Pipeline}
\postsec
\label{sec:biasedpipeline}
For the biased pipeline, the starting point is the unbiased iteration-1 model that was trained for 250k updates (\ie $M^{250k}_{100}$).
This model is finetuned for 20k updates using a \gls{ctc} loss on the labelled data. This is the biasing step. These 20k updates come after the linear output layer is trained for 10k updates itself, as done for the finetuning (see \sectdot{finetuning}) as well\footnote{Finetuning for only 5k updates also gave good results, but training for too long (\eg 80k updates) started to give worse results, which could be due to overfitting to the supervised subset of the data.}. Then the embeddings of the \nth{17} emformer-layer are clustered to obtain the target sequence. The \nth{17} layer was picked as it maximised the label-purity. The more similar the model is to an \gls{asr} model, the closer the ideal layer for the embeddings to the output. 
For the biased training, both K=500 and K=100 are tested. These pre-trained models are $C^{\alpha}_{\beta}$. 

\presec
\subsection{Training Parameters}
\postsec
For the \gls{mppt} of any iteration  (biased or unbiased), around 50\% of the input frames are masked by applying overlapping masks of length 20 frames (equivalent to 200~ms; \cite{hsu2021hubert} used masks length 10 for 50~Hz features \ie also 200~ms). This is done by choosing 4\% of frames as starting frames for the masks of length 20. The classification loss is computed only for the masked frames. The models are trained on 32 GPUs with a batch size of 87.5 seconds per GPU (for 250k updates, this is equivalent to 169 epochs). The learning rate ramps up linearly for the first 8\% of updates and then decays linearly down to 0. The peak learning rate is 5e-4.

\presec
\subsection{Supervised Baselines and Finetuning}\label{sec:finetuning}
\postsec
%
The pre-trained models are finetuned using the \gls{ctc} loss~\cite{Graves2006CTC}. The output units are 5000 sentence pieces~\cite{kudo2018sentencepiece} with \gls{bpe}~\cite{sennrich2015neural} as the segmentation algorithm.
For finetuning, the projection and output layer of the pre-trained models are replaced with an output layer mapping to the 5001 output units (incl. the blank unit). The pre-trained encoder is frozen for the first 10k updates, and the entire model was trained for 100k updates (10k+90k). The peak learning rate was 2.5e-5, which is 40x smaller than for the purely supervised models (1e-3). The purely supervised models have the same model structure as the pre-trained models. All models use SpecAugment~\cite{park2019specaugment} without time-warping. For decoding, the official Librispeech 4-gram language model was used.
\presec
\section{Experimental Results}
\postsec
\label{sec:expres}

\begin{table}[tb]
    \centering
    \begin{tabular}{cccccc}
        \toprule
        \multirow{ 2}{*}{ID} & \multicolumn{2}{c}{Embeddings} & \multirow{ 2}{*}{K} & \multirow{ 2}{*}{T-steps} & \multirow{ 2}{*}{M-acc}\\\cmidrule{2-3}
        & Model & Layer & & & \\\midrule
        $M^{100k}_{100}$ &\multicolumn{2}{c}{39D-MFCC} & 100 & 100k & 48.4\\
        $M^{250k}_{100}$ & \multicolumn{2}{c}{39D-MFCC} & 100 & 250k & 49.5\\\midrule
        $H^{100k}_{500}$ & $M^{250k}_{100}$ \phantom{+  ft.}& 10 & 500 & 100k & 58.3\\
        $H^{250k}_{500}$ & $M^{250k}_{100}$ \phantom{+  ft.}& 10 & 500 & 250k & 59.1\\
        $H^{400k}_{500}$ & $M^{250k}_{100}$ \phantom{+  ft.}& 10 & 500 & 400k & 61.0\\\midrule
        $C^{100k}_{500}$ & $M^{250k}_{100}$ +  ft. & 17 & 500 & 100k & 67.7\\
        $C^{250k}_{500}$ & $M^{250k}_{100}$ +  ft. & 17 & 500 & 250k & 70.1\\
        $C^{400k}_{500}$ & $M^{250k}_{100}$ +  ft. & 17 & 500 & 400k & 71.5\\\midrule
        $C^{100k}_{100}$ & $M^{250k}_{100}$ +  ft. & 17 & 100 & 100k & 77.0\\
        $C^{250k}_{100}$ & $M^{250k}_{100}$ +  ft. & 17 & 100 & 250k & 79.1\\
        \bottomrule
    \end{tabular}
    \vspace*{0.8em}
    \caption{\Acrlong{mppt}. ID is the ID of the model that was trained. Embeddings are the vectors clustered into $K$ clusters to obtain the training sequence. M-acc is the model's accuracy on the masked frames on the validation set. T-steps is the number of training updates. '+ft.' indicates the biasing step.}
    \label{tab:huberttrain}
\end{table}
\begin{table}[tb]
    \centering
    \begin{tabular}{cccccH}
        \toprule \multicolumn{2}{c}{Embeddings} & \multirow{ 2}{*}{K}& cluster- & label- & H\\\cmidrule{1-2}
        Model & Layer & & purity & purity\\\midrule
        
        \multicolumn{2}{c}{39D-MFCC} & 100 & 0.074 & 0.064 & 0.111\\
        $M^{250k}_{100}$ \phantom{+ ft.} & 10 & 500 & 0.079 & 0.162 & 0.323\\
        $M^{250k}_{100}$ \phantom{+ ft.}& 10 & 100 & 0.152 & 0.114 & 0.234\\
        $M^{250k}_{100}$ + ft. & 17 & 500 & 0.299 & 0.197 & 0.370\\
        $M^{250k}_{100}$ + ft. & 17 & 100 & 0.330 & 0.194 & 0.360\\
        \bottomrule
    \end{tabular}
    \vspace*{0.8em}
    \caption{Label purity is the label prediction accuracy if cluster is known. Cluster purity is the cluster prediction accuracy if label is known. 
    Labels are a set of 870 clustered bi-characters.}
    \label{tab:labelstats}
\end{table}
\begin{table}[tb]
    \centering
    \begin{tabular}{rrrr}
        \toprule
        Data & Init. & test-clean & test-other\\\midrule
        100h & - & 8.68& 17.76\\\midrule
        100h & $M^{100k}_{100}$ & 6.98& 14.01\\
        100h & $H^{100k}_{500}$ & 6.52 & 13.00\\
        100h & $C^{100k}_{500}$ & 5.28 & 10.39\\
        100h & $C^{100k}_{100}$ & 5.06 & 9.91\\
        100h & $S^{100k}_{100}$ & 4.64 & 8.93\\
        100h & -Conv1D & 4.95 & 10.30\\
        100h & +K=2000 & 4.81 & 9.93 \\\midrule
        100h & $M^{250k}_{100}$ & 6.26 & 12.68\\
        100h & $H^{250k}_{500}$ & 5.27 & 10.72\\
        100h & $C^{250k}_{500}$ & 4.67 & 9.60\\
        100h & $C^{250k}_{100}$ & 4.48 & 9.06\\
        100h & $S^{250k}_{100}$ & 4.50 & 8.75\\\midrule
        100h & $H^{400k}_{500}$ & 5.02 & 10.23\\\midrule
        960h & - & 3.26& 7.37\\
        \bottomrule
    \end{tabular}
    \vspace*{0.8em}
    \caption{
    Results on Librispeech test sets after CTC finetuning. Data is hours of finetuning data. Init. is the model used for initialisation.}
    \label{tab:WERCTC}
\end{table}

The masked prediction accuracies are given in \tbl{huberttrain}. 
The label-statistics are given in \tbl{labelstats} and \glspl{wer} are given in \tbl{WERCTC}.
\glspl{wer} in the text will refer to test-other except when explicitly mentioned.

\presec
\subsection{Baselines}
\postsec
The \glspl{wer} of the baseline 100-hour and 960-hour purely-supervised models are 17.76\% and 7.37\%, respectively. The unbiased models $M^{250k}_{100}$ and $H^{250k}_{500}$ achieve \glspl{wer} of 12.68\% and 10.72\%. 
Thus, two iterations of unbiased pre-training reduce the \gls{wer} by 39.6\% over the supervised baseline and ``recover" 67.8\% of the reduction in \gls{wer} obtained from 860 hours of labelled data (17.76\%$\,\to\,$7.37\%).
Training for the second iteration for 100k updates (\ie initialised from $H^{100k}_{500}$) only yields a 26.8\% reduction (13.00\% \gls{wer}).

The clusters derived from layer 10 of $M^{250k}_{100}$ give better predictions of the bi-character labels than the clusters from the MFCC features (label-purity of 0.114 vs 0.064) for K=100. 

Looking at \tbl{huberttrain}, it can be seen that even with the increase in the number of clusters from 100 to 500, the masked-prediction accuracy is higher for $H^{250k}_{500}$ than for $M^{250k}_{100}$ (59.1\% vs 49.5\%). This is likely due to the tokenisation being more consistent.

\presec
\subsection{Biased HuBERT training}
\postsec
To recall, before clustering, the iteration-1 model, $M^{250k}_{100}$, is finetuned for 20k update steps. The embeddings of the \nth{17} layer are clustered. For K=500, the purity statistics improve significantly (see \tbl{labelstats}) over the unbiased models. The cluster-purity increases from 0.079 to 0.299 and the label-purity from 0.162 to 0.194. The strong increase in cluster-purity confirms the idea that without supervision, many cluster tokens represent the same phonetic unit but in different acoustic contexts (speaker, channel, noise). Part of the challenge of acoustic modelling is to be invariant to changes in context. Thus, a higher cluster-purity is desired.
Another effect of the biasing is that K=100 seems to be large enough to model the bi-character labels. For the unbiased embeddings, the label-purity drops from 0.162 down to 0.114 when changing K=500 down to K=100. For the biased embeddings, only a small drop from 0.197 to 0.194 is observed. This observation motivated the decision to try K=100 for the biased second iteration of training and not only K=500 as commonly used~\cite{hsu2021hubert,chen2022wavlm,wang2022supervision}.

From \tbl{huberttrain}, it can be observed that the biased labels are much cleaner and more predictable. The masked prediction accuracy, after 250k training steps, increases from 59.1\% to 70.1\% (K=500 for both). Using K=100 has a masked prediction accuracy of 79.1\%.

For 250k updates, the biasing model (init. from $C_{500}^{250k}$ outperforms the unbiased model (init. from $H_{500}^{250k}$) by 10.4\% (10.72\% \gls{wer} vs. 9.60\% \gls{wer}). Using K=100 further reduces the \gls{wer} to 9.06\% (15.5\% rel. \gls{wer} reduction). Therefore the biased training is able to recover 83.7\% of the reduction in \gls{wer} that could be obtained by using 860 hours of labelled data in comparison to the 67.8\% achieved by the unbiased pre-training. 

The improvements due to the biasing are even stronger when training for only 100k updates. The relative \gls{wer} reduction over the unbiased baseline is 20.1\% for K=500 and 23.8\% for K=100. This demonstrates that biased \gls{ssl} improves not only the final performance but also the convergence speed. The main reason for this is likely the improved consistency of the label sequences and the thus increase in the predictability of the cluster-tokens, which can be seen from the increase in masked-prediction accuracy in \tbl{huberttrain}. Label noise in supervised learning, which can be seen as analogous to less consistency in the tokenisation obtained from clustering, hinders convergence as shown by~\cite{zhang2017understanding}.

For an iteration-3 model where the label sequence is obtained by clustering the embeddings of $C^{250k}_{100}$, the convergence speed increases even more (this model is called $S^{\alpha}_{100}$). Even though \gls{wer} for 250k updates is only 3.4\% lower (8.75\% vs 9.06\%) than for the biased iteration-2 model (both with K=100), for 100k updates, the WER is 9.9\% lower (8.93\% vs 9.91\%). This again reinforces the connection between how clean the label-tokens are and the convergence speed.

\presec
\subsection{Low-footprint streaming models}
\postsec
The labels used for the biased iteration-3 models ($S^{\alpha}_{100}$) are treated as very good acoustic units and used for the CE-style pre-training for which the loss comes equally from the masked and the unmasked frames.
In the first experiment, this style of training is used while at the same time removing the 1D-Convolutional layer. This increases the \gls{wer} from 8.93 to 10.30\%. Using a larger token set of K=2000 reduces the \gls{wer} down to 9.93\%. Here, a larger token set helps because another interpretation of this training method is that the model is distilled from $C^{250k}_{100}$. A larger token set gives more information about $C^{250k}_{100}$.

For a low-footprint streaming model (with an algorithmic latency of 120 ms), the 100-hour supervised \gls{wer} is much worse at 12.45\%/24.07\% \gls{wer} on test-clean/test-other. Using our proposed style of training, the \gls{wer} is reduced to {6.13\%/13.34\%} for the same low-footprint streaming model. A reduction in \gls{wer} of 50.8\%/44.6\%.
\presec
\section{Conclusions}
\postsec
\label{sec:conclusions}
Biased self-supervised learning is a framework for semi-supervised learning of speech representations by biasing the labels of the masked-prediction pre-training towards \gls{asr}. Our experiments show large reductions in \gls{wer} over the unbiased training mode. Furthermore, a substantial improvement in training speed can be seen such that biased training for 100k updates outperforms unbiased training for 400k updates. This achievement also highlighted the strong connection between label noise and convergence speed, which should be kept in mind for developing future self-supervised learning algorithms. Furthermore, we showed the use of biased \gls{ssl} for small-footprint streaming \gls{asr} models. 

\vfill\pagebreak

\section{REFERENCES}
\label{sec:refs}

\begingroup
\renewcommand{\section}[2]{}
\bibliographystyle{IEEEbib}
\bibliography{refs}
\endgroup
\end{document}